\documentclass[fleqn,10pt]{wlscirep}
\usepackage[utf8]{inputenc}
\usepackage[T1]{fontenc}
\title{Extrapolating tipping points and simulating non-stationary dynamics of complex systems using efficient machine learning}

\author[1,*]{Daniel Köglmayr}
\author[1]{Christoph Räth}

\affil[1]{German Aerospace Center (DLR), Institute for AI Safety and Security, Ulm,
89081, Germany}

\affil[*]{daniel.koeglmayr@dlr.de}

\begin{abstract}
Model-free and data-driven prediction of tipping point transitions in nonlinear dynamical systems is a challenging and outstanding task in complex systems science.
We propose a novel, fully data-driven machine learning algorithm based on next-generation reservoir computing to extrapolate the bifurcation behavior of nonlinear dynamical systems using stationary training data samples. We show that this method can extrapolate tipping point transitions.
Furthermore, it is demonstrated that the trained next-generation reservoir computing architecture  can be used to predict non-stationary dynamics with time-varying bifurcation parameters.  In doing so, post-tipping point dynamics of unseen parameter regions can be simulated.

\end{abstract}
\begin{document}

\flushbottom
\maketitle
%
%
\thispagestyle{empty}

\section*{Introduction}

Small perturbations in a complex system can dramatically change its evolution \cite{lorenz1963deterministic}. A lack of precision in determining the exact state of the system can lead to an amplified lack of certainty about the future behavior of the system. This is the case even when we know its true governing equations and the exact boundary conditions. 
How can we then deal with complex systems where, in addition, we do not know the governing equations and must rely solely on observational data?

In recent years promising and remarkably efficient machine learning methods were proposed that use observational data as training data to autonomously generate a model that can explain the data \cite{brunton2016discovering,ma2022identifying,huang2022sparse}. One prominent example is a recurrent neural network method called reservoir computing \cite{lukovsevivcius2009reservoir,jaeger2004harnessing} (RC). A reservoir computer creates a high-dimensional nonlinear representation of the observed dynamical system and synchronizes it with the corresponding input data. The synchronized representation is then trained on the desired output target so that the reservoir computer becomes an autonomous dynamical system whose output dynamics resemble that of the analyzed system. This way, it can achieve cutting-edge performances in predicting short-and long-term behavior of chaotic systems and outperforms other machine learning approaches like LSTMs or DNNs\cite{bompas2020accuracy,chattopadhyay2020data}. 
In September 2021, Gauthier et al. published the next-generation reservoir computing architecture (NG-RC), highlighting its lack of randomness, the fewer hyperparameters, the smaller amount of required training data, and its performance gain in speed compared to the traditional approach \cite{gauthier2021next}.
In traditional reservoir computing, randomly initialized matrices are used to feed the input variables of the dynamical system into a high-dimensional state space that is nonlinearized by applying a nonlinear activation function. The NG-RC uses a library of unique polynomials of time-shifted input variables to achieve a nonlinear dimensionality expansion. In both cases, the resulting state space is consistently trained on the desired output target using ridge regression to become an autonomous dynamical system.
Both methods can generally be deployed with small state spaces, which, combined with the computational cheap regression, lead to highly efficient algorithms. 

So far, these algorithms have been used mainly for analyzing $stationary$ dynamical systems, where the boundary conditions of the system are assumed to be fixed, i.e., time-independent. In this case, the qualitative behavior of the system, such as periodicity or chaoticity, remains the same over time. However, in most real-world systems, the boundary conditions can change over time, possibly leading to a qualitative change in the behavior of the system, e.g., from stable periodicity to chaos or from chaos to system collapse. These systems are called $non$-$stationary$ dynamical systems, and the boundary condition under which the system undergoes such a critical transition is termed $tipping$ $point$. Extrapolating tipping points is of great interest in many scientific fields.
A prominent example is the evolution of the climate system influenced by atmospheric greenhouse gas concentrations, for which several tipping points are predicted \cite{lenton2008tipping}. Irrgang et al. surveyed the role of artificial intelligence for earth system modeling. They highlighted the concern that current classic earth system models might not be capable of predicting future abrupt climate changes \cite{irrgang2021towards}. Hence, data-driven methods that capture the underlying physics seem suitable to augment classic models. 

Reservoir computing-based methods for analyzing non-stationary dynamical systems and, thus, for the possible data-driven extrapolation of tipping points are still in their early stages. Two different approaches are emerging in the current literature, which mainly differs from each other in the form of their training data.
One approach directly uses non-stationary time series to train the reservoir computer \cite{lim2020predicting,patel2021using,patel2022using}.
The other approach uses several stationary data samples from different boundary conditions, i.e., bifurcation parameters, to train the reservoir computer \cite{kim2021teaching,kong2021machine,kong2023reservoir}.
The latter uses the multifunctional capabilities of reservoir computing, which are complemented by an additional parameter channel to test for new an unseen parameter regions.
It has been shown that a reservoir computer can be optimized to predict several different dynamical systems with a single trained architecture 
\cite{flynn2021multifunctionality,flynn2021symmetry,flynn2022exploring,herteux2020breaking,lu2020invertible}. 
Kim et al. showed that reservoirs can learn with this approach and the additional parameter channel to interpolate and extrapolate translations, linear transformations, and bifurcations of the Lorenz attractor \cite{kim2021teaching}. In \cite{kong2021machine}, Kong et al. statistically evaluated the parametrization of a system collapse, or global bifurcations, of a chaotic food chain model and a generic power system model.
Kong et al. were also able to reconstruct bifurcation diagrams of driven chaotic systems \cite{kong2023reservoir}.

In this work a framework for parameter-aware next-generation reservoir computing is developed.
By means of the examples used in \cite{kong2021machine}, the functionality of the developed method is demonstrated and it is shown that the method is capable of accurately reconstructing bifurcation diagrams and simulating non-stationary dynamics, even in situations where the data is limited and the parameterization of the training data is far from the global bifurcation of interest.

\section*{Results}

Recently, a new type of RC called next-generation reservoir computing (NG-RC) has been introduced for the analysis of dynamical systems.
In its functional core, the algorithm first collects the time-shifted input variables of the time series data to be analyzed into a vector. In a second step, each unique polynomial combination of certain orders of the entries in the previously collected vector is determined and appended. In this way, the feature vector is created. During training, the linear mapping of the feature vector to the corresponding next time series data point is optimized using ridge regression. Due to this minimal architecture, the NG-RC features excellent speed and lacks any randomness. Besides these operational advantages, it has been shown in several publications that NG-RC requires significantly less training data than the already data sparing traditional reservoir computer \cite{gauthier2021next,barbosa2022learning,gauthier2022learning, haluszczynski2023controlling}, which makes NG-RC a highly efficient method for analyzing and predicting dynamical systems.

The algorithm proposed in this paper models an additional input channel for a bifurcation parameter into the NG-RC architecture by adding the parameter times a scaling parameter to each entry of the feature vector (see Methods). This allows the algorithm to learn a dynamical system also in terms of its bifurcation parameter. After training, the parameter can be varied so that the prediction of the algorithm can be tested for unseen parameter regions. 
The parameter-aware next-generation reservoir computing architecture is applied below to two systems of ordinary differential equations, a generic power system model \cite{dobson1989towards} and a chaotic food chain model \cite{mccann1994nonlinear}. Both systems were examples for statistical evaluation of tipping points using traditional reservoir computing \cite{kong2021machine}, moreover, their equations contain terms that cannot be directly represented by the polynomial structure of the feature vector, making them informative test systems for the parameter-aware NG-RC.

For the generic power system model, the NG-RC architecture is used to predict the bifurcation diagram and to extrapolate the tipping point. To evaluate its prediction quality, the largest Lyapunov exponents are compared with those of the model equations. The Lyapunov exponent is a measure of the long-term statistical behavior, or statistical climate, of a time series and indicates how chaotic or periodic a time series is (see Methods). It is also demonstrated that the correct choice of the scaling parameter is important and affects the quality of the prediction. For the chaotic food chain model, the influence of the scaling parameter on the extrapolation is further investigated. For that, 15 bifurcation diagrams predicted by the same NG-RC architecture are shown, which only differ in a slightly different scaling parameter. 
Furthermore, it is shown that the trained NG-RC can be used to simulate non-stationary dynamics in unseen parameter regions, capturing the main behavior of the dynamics even after passing through a tipping point.

\subsection*{Power system model}\label{Power system model}

\begin{figure*}[!htb]
\includegraphics[width=\linewidth]{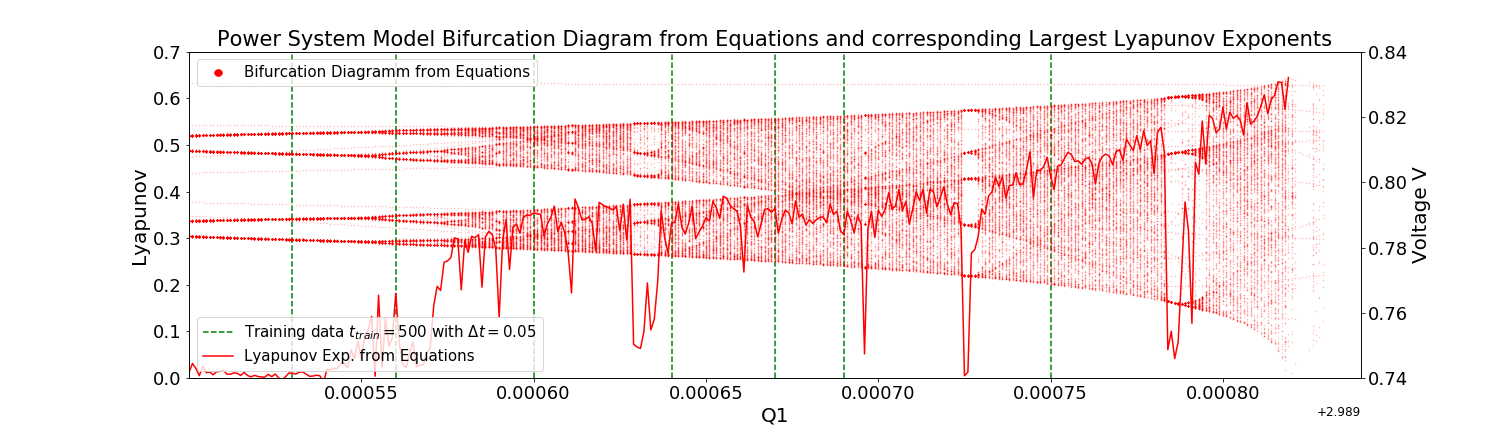}
\includegraphics[width=\linewidth]{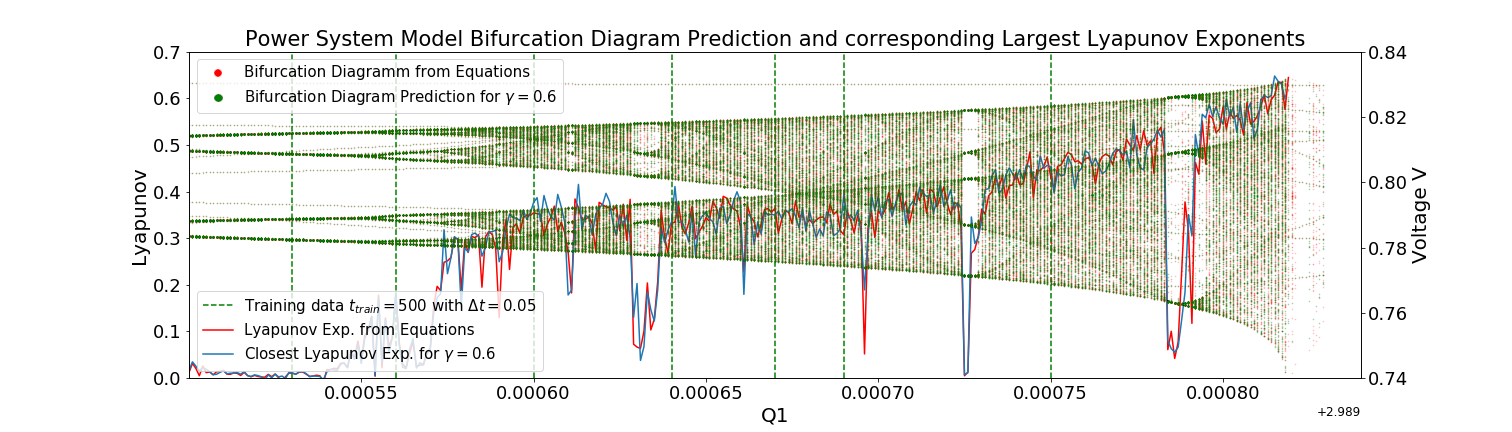}
\includegraphics[width=\linewidth]{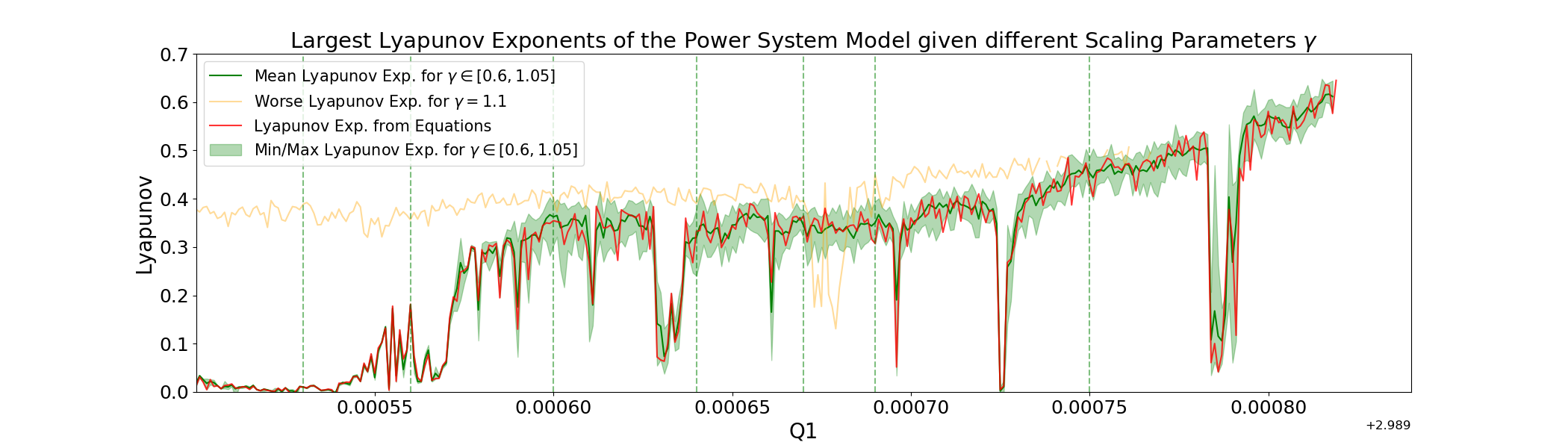}
\caption{\label{fig:Volltage_Collapse}
Upper Plot: The bifurcation diagram of the power system model with a voltage collapse at $Q_{1c}=2.989 + 0.000819$.
Middle Plot: The predicted bifurcation diagram of the power system model using the proposed NG-RC architecture with a scaling parameter  of $\gamma=0.6$ is scattered in green. The architecture was trained with seven stationary training data samples. The corresponding bifurcation parameters are highlighted as green dashed lines with 10000 training data points each. For visual comparison, the bifurcation diagram of the model equations is scattered in red and also shown in the upper plot. The respective largest Lyapunov exponents of the two systems are plotted for the quantitative comparison. Both show systematic similarities. Lower Plot: The NG-RC architecture is tested for different scaling parameters. The enclosed area of minimum and maximum predicted Lyapunov exponents for scaling parameters in $\gamma \in [0.6,1.05]$ is highlighted in light green. All capture the systematic behaviors of the model. Above a scaling parameter of $\gamma=1.1$, the NG-RC architecture loses its predictive power. 
}
\end{figure*}

Dobson and Chiang formulated a set of generic equations to model the collapse of electrical power systems.  This can be caused by the dynamic response of the system to disturbances, which may lead to a progressive drop in voltage, causing what is known as a "voltage collapse" or blackout \cite{dobson1989towards}. In the upper plot of Fig. \ref{fig:Volltage_Collapse}, the bifurcation diagram of the generic equations of the power system model is scattered in red. The corresponding Lyapunov exponents are plotted to measure the dynamic behavior of the system.  It evolves from a periodic dynamic to a chaotic one for increasing bifurcation parameters. In some areas, it shows periodic windows. The system collapses at the critical bifurcation parameter $Q_{1c}=2.989820$, and the total voltage drops to zero.
The presented NG-RC architecture aims to reconstruct the bifurcation diagram with matching Lyapunov exponents.  In this example, seven training data samples are taken from different and widely separated regions of the bifurcation diagram to be analyzed. The bifurcation parameter of these are highlighted as vertical green dashed lines.

\subsubsection*{Results}

The parameter-aware NG-RC architecture was applied and tested with different scaling parameters.
The reconstructed bifurcation diagram of the best performing architecture is scattered in the middle plot of Fig. \ref{fig:Volltage_Collapse} in green, and its corresponding Lyapunov exponents are plotted in blue. It captures the main dynamical behaviors of the model. Between the area of training data samples, the architecture interpolates the periodic windows even though none of the samples were set in a similar region. In extrapolating the dynamics, the architecture captures the periodic window starting at $Q_1=2.989784$ and predicts the system collapse at $Q_{1c}=2.989819$.
The dynamical properties of the model were successfully captured for a set of scaling parameters in the range of  $\gamma \in [0.6,1,05]$. The corresponding minimum and maximum Lyapunov exponents of these parameters are plotted in the lower plot of Fig. \ref{fig:Volltage_Collapse}. The Lyapunov exponents of the model equations lie well in between this region.
The Lyapunov exponents for $\gamma=1.1$ are plotted in yellow, showing that the NG-RC architecture could not predict the dynamical properties of the model. 
For some parameters of the training data, the Lyapunov exponents of the predicted dynamics differ strongly from those of the training data, which allows for direct validation of prediction performance given the applied scaling parameter.
This makes the introduced scaling parameter a functional new hyperparameter, which is worthwhile tuning in this setup. Its functionality is further investigated in the next example.

\begin{figure*}[!htb]
\includegraphics[width=\linewidth]{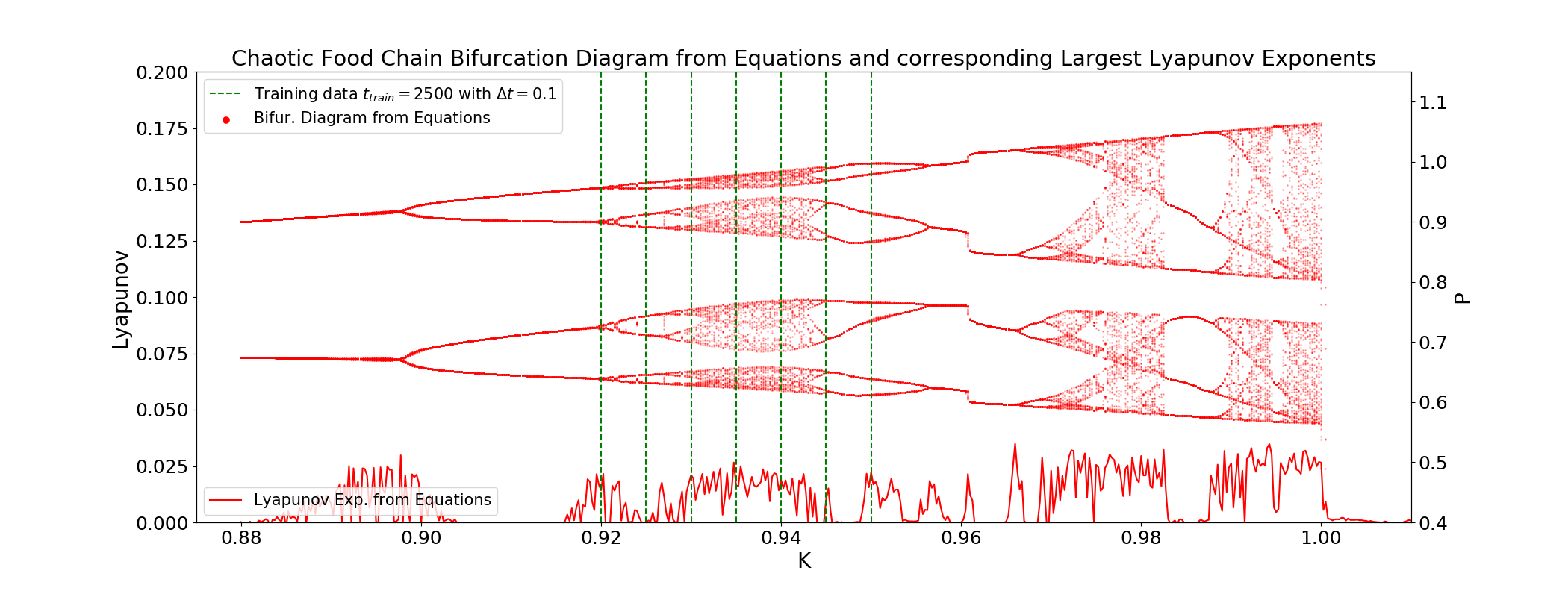}
\includegraphics[width=\linewidth]{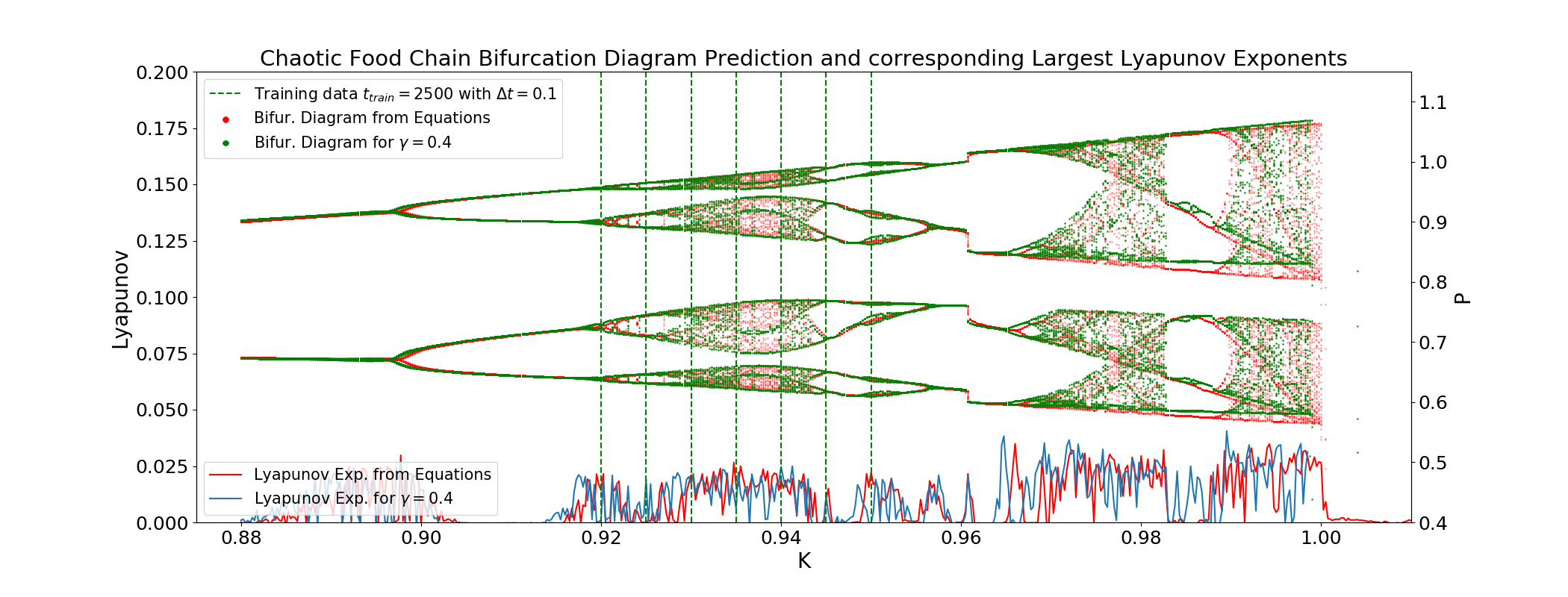}
\caption{\label{fig:CFC}
Upper Plot: The bifurcation diagram of the chaotic food chain model.
Lower Plot: Testing the extrapolation capabilities through a reduced parameterization range of training data samples. The predicted bifurcation diagram of the chaotic food chain model using the proposed NG-RC architecture with a scaling parameter  of $\gamma=0.4$ is scattered in green. The architecture was trained with seven stationary training data samples and with 25000 training data points each. The bifurcation diagram of the model equations is scattered in red. The respective largest Lyapunov exponents of the two systems are plotted for the quantitative comparison. The predicted diagram shows systematic similarities that are shifted on the $K$-axis the further away they are from the parameterization of the training data.}
\end{figure*}

\newpage
\subsection*{Chaotic food chain model}

McCann and Yodzis showed that ecosystem behavior, when it transitions to chaotic transient dynamics, can cause sudden and unpredictable disappearance of populations. Due to the realistic and nonlinear functional response properties of productive environments, sudden and unexpected jumps to other dynamical population density attractors may occur, potentially causing the disappearance of a population \cite{mccann1994nonlinear}. To model this, they used a three-species food chain model with a resource density $R$, a consumer density $C$, and  a predator density $P$. The resource-carrying capacity $K$ of the environment is taken as the bifurcation parameter. The bifurcation diagram of the model equations is scattered in Fig. \ref{fig:CFC} in red and with a larger $K$ space in Fig. \ref{fig:CFC_scaling_plot}. 
This system of equations shows rich bifurcation structures.
When the resource-carrying capacity $K$ reaches a critical value of $K_{c_1}=1.00050$, the chaotic oscillating predator density $P$ suddenly drops to $0$, and the predator population disappears. For $K_{c_2}=1.04$, this density reappears in a reverse manner. Both system behaviors are global bifurcations. Notably, there is another one at $K_{c_3}=0.96075$, where the predator density performs a sudden jump.
This time the NG-RC architecture aims to reconstruct
the bifurcation diagram given stationary training data samples, which are narrowed down to a more minor part of the bifurcation diagram compared to those taken in the previous example. The performance of the extrapolation of tipping points is evaluated regarding the introduced scaling parameter. Non-stationary dynamics are simulated, passing through the tipping point $K_{c_3}$.

\subsubsection*{Results}

\begin{figure*}[!htb]
\centering
\includegraphics[width=0.82\linewidth]{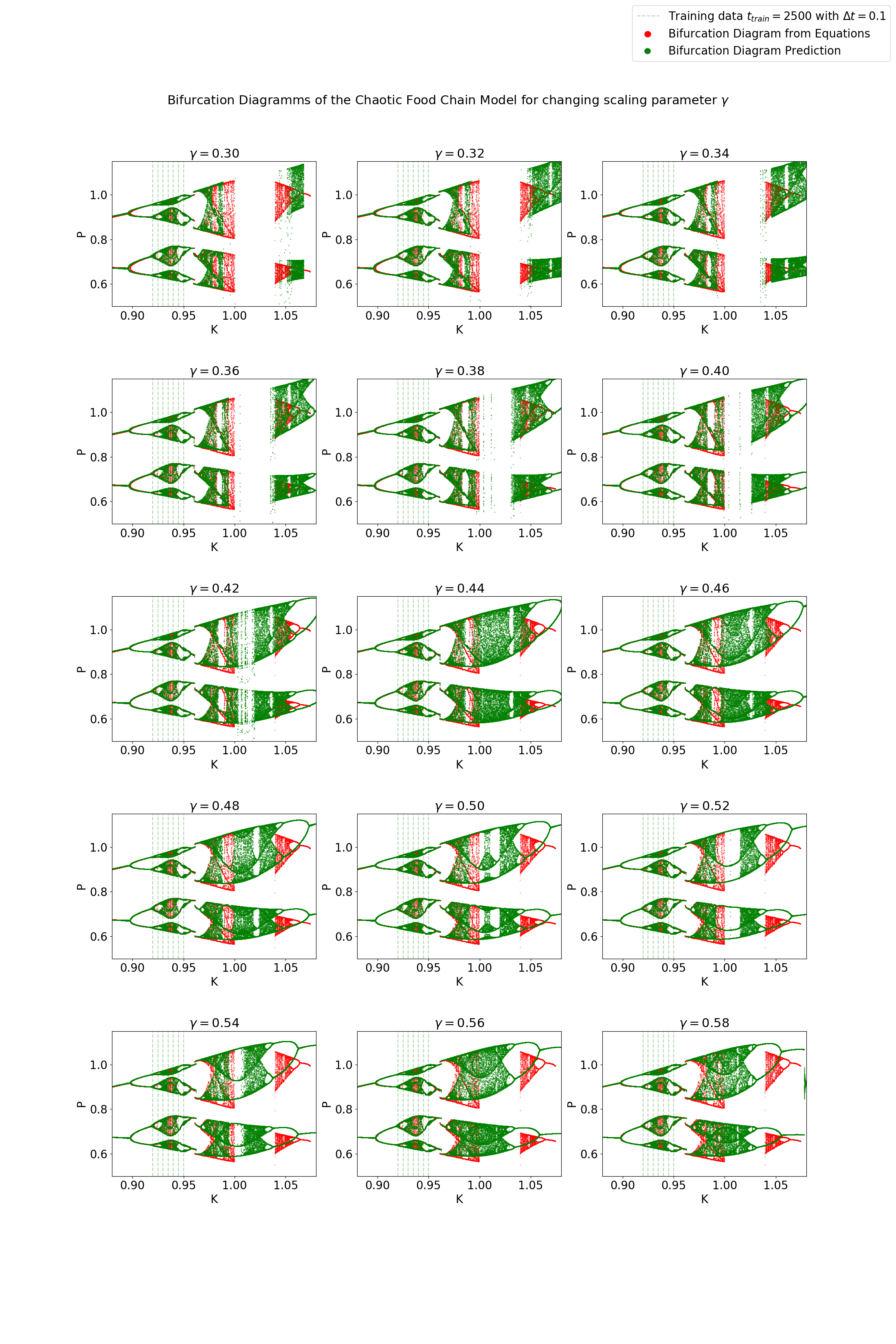}

\caption{\label{fig:CFC_scaling_plot}The NG-RC architecture was tested for different scaling parameters on the chaotic food chain model. This illustrates the effect of the newly introduced hyperparameter on the system behavior of the learned NG-RC dynamic and the extrapolation capabilities of tipping points.}
\end{figure*}

\begin{figure*}[!htb]
\includegraphics[width=\linewidth]{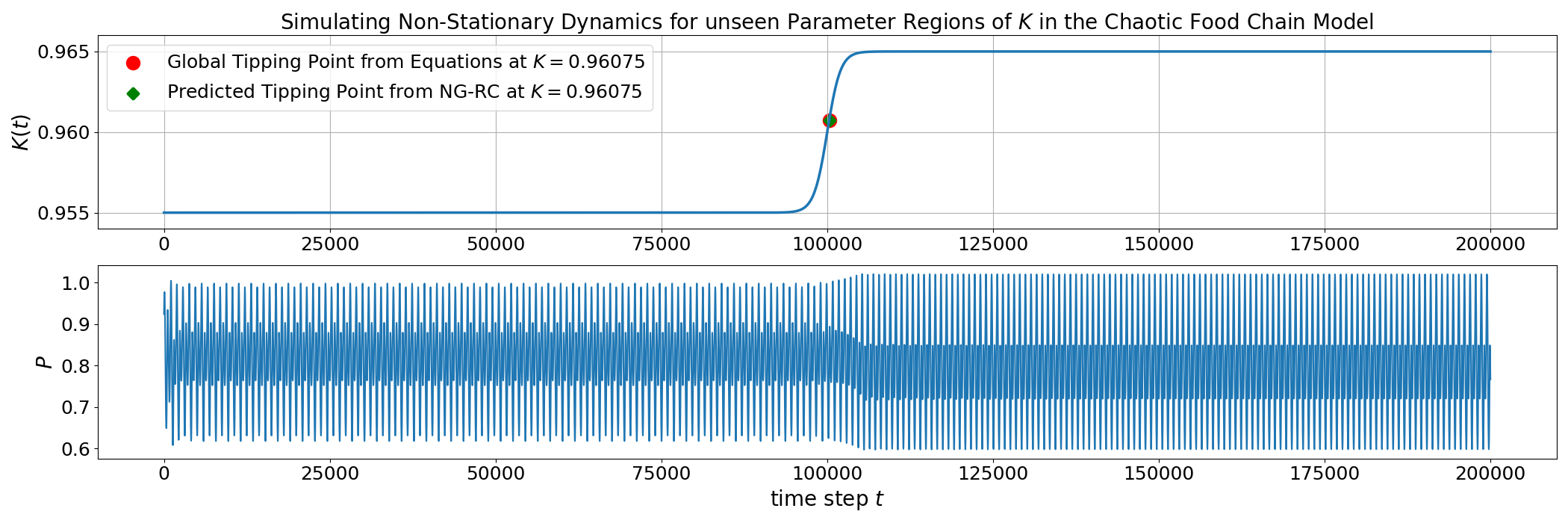}
\caption{\label{fig:cfc_switch}Simulating non-stationary dynamics of the chaotic food chain model with the trained NG-RC architecture of Eq. \ref{CFC_good} for unseen parameter regions. The switch of the bifurcation parameter $K$ across the tipping point at $K_{c_3}=0.96075$ shows that the simulated trajectory captures the expected change in behavior (see Fig. \ref{fig:CFC}).}
\end{figure*}

\begin{figure*}[!htb]
\includegraphics[width=\linewidth]{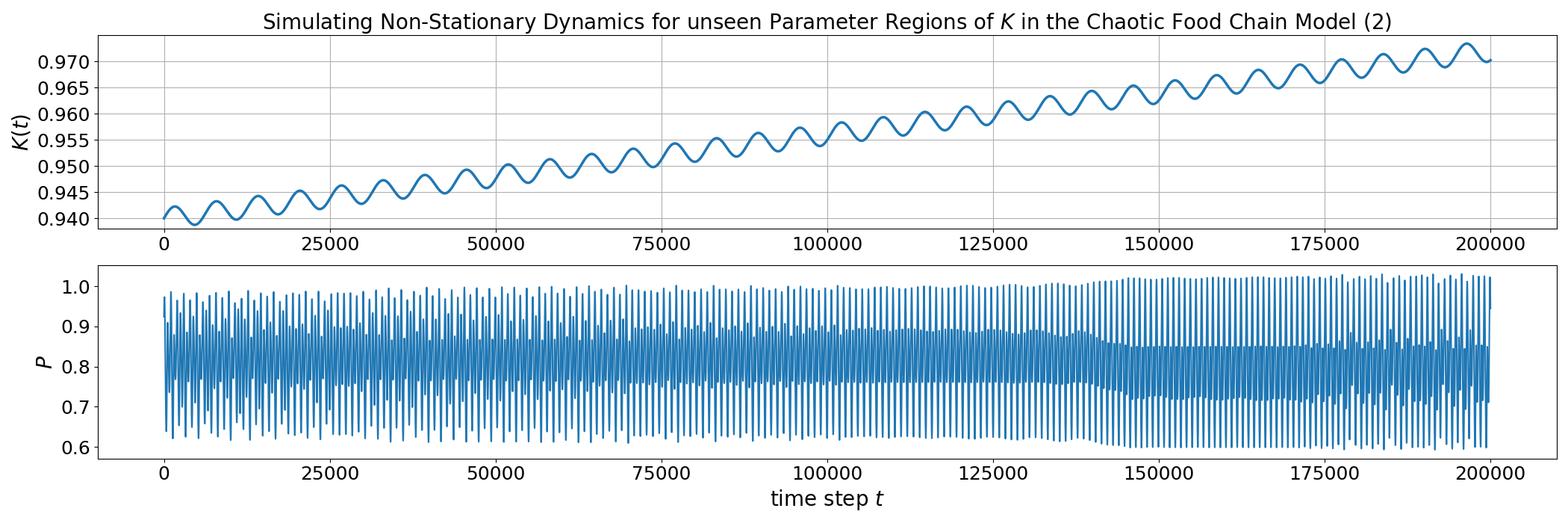}
\caption{\label{fig:cfc_sin} As another example, the trained NG-RC architecture of Eq. \ref{CFC_good} is used to simulate non-stationary dynamics of the chaotic food chain model for a more complicated sinusoidal and linearly increasing bifurcation parameter $K$. The simulated trajectory captures the qualitative behaviors of the respective parameter regions (see Fig. \ref{fig:CFC}). }
\end{figure*}

The prediction performance of the NG-RC architecture was investigated for different scaling parameters. The best performing architecture  with $\gamma=0.4$ is scattered in Fig. \ref{fig:CFC} in green. The nearest tipping point from the parameterization of the training data at $K_{c_3}=0.96075$ and the subsequent transition from chaoticity to periodicity at $K=0.983$ are accurately predicted.
The tipping point at $K_{c_1}=1.00050$ was predicted with $K=0.99875$. The bifurcation that is farthest away, $K_{c_2}=1.04$, was extrapolated with $K=1.027$ (see Fig. \ref{fig:CFC_scaling_plot}). A qualitative difference between the prediction and the model is that the predicted trajectory after the tipping point at $K=0.99875$ goes to minus infinity, whereas the real one goes to zero. Clear topological differences can be seen between the training data samples at $K=0.935$ and $K=0.94$. This area incorrectly shows the properties of a periodic window. This also applies to the interpolation for different scaling parameters.  Thus, the effect of the scaling parameter on the interpolation capabilities between the parameter space of training data samples is limited. As expected, accurate extrapolation of possible tipping points becomes more difficult the further they are from the parameterization of the training data.
Looking at the evolution of the bifurcation diagrams for different scaling parameters in Fig. \ref{fig:CFC_scaling_plot} was instructive to see the influence of the scaling parameter on the extrapolation capability. From $\gamma=0.3$ on to $\gamma=0.4$, the increasing scaling parameter stretches the bifurcation topology  in the range of $K \in [0.95,1]$. Interestingly, there is a qualitative change in the bifurcation diagram when the scaling parameter stretches it over the actual tipping point at $K_{c_1}=1.00050$. The tipping point prediction is lost, and the NG-RC transforms the two parts of the bifurcation diagram into a continuous one. This behavior generally allows practical parameter tuning of the scaling parameter by introducing validation data to determine the necessary degree of stretching.
Moreover, the trained NG-RC architecture  for $\gamma=0.4$ is used to simulate non-stationary dynamics using Eq. \ref{non-stationary equation}. In Fig. \ref{fig:cfc_switch}, the bifurcation parameter switches from $K=0.955$ to $K=0.965$ over the tipping point at $K_{c_3}=0.96075$. The predicted trajectory captures this transition. Using the identical trained NG-RC architecture, a sinusoidal and linearly increasing function of $K$ is taken as another example. The result is shown in Fig. \ref{fig:cfc_sin}. The dominant dynamical behaviors concerning the bifurcation parameter regions are captured in the prediction.
These examples illustrate the applicability of this architecture, which enables the simulations of non-stationary dynamics as a function of a time-varying bifurcation parameter.

\newpage
\section*{Discussion}

A machine learning method based on parameter-aware next-generation reservoir computing was presented to investigate the bifurcation behavior of dynamical systems. It was shown that tipping points can be extrapolated. Moreover, the trained architecture can be used to simulate non-stationary dynamics and, with that, also, post-tipping point dynamics. The success of reservoir computing and next-generation reservoir computing relies on optimizing them into dynamical systems whose dynamics resemble that of the analyzed system. The presented implementation of the bifurcation parameter provided a functional input channel that allowed the investigation of the system dynamics in unseen parameter regions.
It is noteworthy that, on the one hand, this method can capture high sensitivities of the analyzed dynamical system to the bifurcation parameter. This was shown in the generic power system model, where the sixth decimal place of the bifurcation parameter partly determines the dynamic behavior. On the other hand, this method integrated the bifurcation parameter, implemented by adding it times a scaling parameter to the feature space of NG-RC, so that the optimized NG-RC architecture was able to simulate and predict dynamics where the bifurcation parameter appears as an inverse parameter in the governing equation. This generally extends the applicability of this approach and was shown in the chaotic food chain example.
Although both system equations contain terms that cannot be directly represented by the polynomial structure of the feature vector, the proposed NG-RC architecture was able to interpolate and extrapolate the system behaviors, which is another plus for its applicability.

So far, there are few publications in which reservoir computing methods are used to determine bifurcation diagrams of dynamical systems and their tipping points. In Kim et al.\cite{kim2021teaching}, parameter-aware reservoir computing was used to accurately extrapolate the period doubling bifurcations of the Lorenz system around $\rho \approx 100$. For this, 4 training samples with 250000 training steps and 50000 synchronization steps were used, resulting in 1200000 data points.  In Kong et al.\cite{kong2023reservoir}, the bifurcation diagram of a driven Lorenz-96 system was predicted with parameter-aware RC. This was done using 4 training samples with 140000 training steps and 800 synchronization steps each, resulting in 563000 data points. 
In the results presented here, the parameter-aware NG-RC required 70014 data points to train the architecture on the power system model and 175112 data points for the chaotic food chain model. A general statement that NG-RC requires significantly less training data than traditional RC, even in the case of parameter-aware extrapolation, would be overstated due to the lack of a direct comparison of the two methods. However, the results presented here provide a first tendency that the required training data can significantly be reduced.
In terms of setting up a working architecture, the parameter-aware RC approach has eight tunable hyperparameters \cite{kong2023reservoir}, while the proposed NG-RC architecture has six, most of which are far less comprehensive to optimize. In addition, the NG-RC works completely without randomness. Instead of random matrices, the polynomial architecture generally ensures higher interpretability and, together with the previously mentioned points, a more direct setup to deploy a working architecture without stochastic realizations of the reservoir system. 
In the context of this work, the NG-RC architectures were not extensively optimized nor comprehensively investigated concerning the minimal required training data. However, if the tendency holds, further applications emerge. Since most real-world dynamical systems are of non-stationary nature, the less training data needed, the better non-stationary data samples can be approximated as stationary data. Which can improve the prediction of tipping points based on non-stationary time series data. Consequently, the here proposed parameter-aware NG-RC is an efficient, model-free, and data-driven method for extrapolating the behavior of dynamical systems and simulating non-stationary dynamics.

\section*{Methods}

The method presented here is based on next-generation reservoir computing. Its architecture is extended by an input channel for a bifurcation parameter of a dynamical system. Therefore, the parameter is added to the NG-RC feature vector as a product with a scaling parameter to each element of the feature vector. The new feature vector is then extended with orders of itself. These steps are presented in detail below. A condensed mathematical description of the NG-RC is used, so that the applied architecture and its hyperparameters can be written as one equation. For the power system model results the architecture with its hyperparameters is described in Eq. \ref{PSM_good} and for the chaotic food chain model in Eq. \ref{CFC_good}.

\subsection*{Next-generation reservoir computing}

The d-dimensional data points $\mathbf{x} \in \mathbb{R}^d$ of the input data $\mathbf{X}=(\mathbf{x}_0,....,\mathbf{x}_n)$ are transformed with a polynomial multiplication dictionary $\mathbf{P}$ into a higher dimensional state space. The unique polynomials of certain orders $O$, included in $\mathbf{P}^{[O]}$, are denoted by an index. For illustration purposes, we consider a two-dimensional input data point 
$\mathbf{x}_i=(x_{i,1}, x_{i,2})^T$ and transform it with the unique polynomials of order 1 and 2,

\begin{equation}
    \mathbf{P}^{[1,2]}(\mathbf{x}_i)=
    \begin{pmatrix}
    x_{i,1}\\[\jot]x_{i,2}\\[\jot]x_{i,1}^2\\[\jot]x_{i,2}^2\\[\jot]x_{i,1}x_{i,2}
    \end{pmatrix}.\
	\label{eq:quadratic}
\end{equation}
Further, Gauthier et al. introduced a time shift expansion $\mathbf{L}_k^s$ of the input data.
The $k$ value indicates the number of past data points with which the current data point is concatenated. The $s$ value indicates how far these points are separated in time.
Following the previous example

\begin{equation}
    \mathbf{P}^{[1,2]}(\mathbf{L}^{s=1}_{k=2}(\mathbf{x}_i))=
    \mathbf{P}^{[1,2]}(
    \begin{pmatrix}
    x_{i,1}\\[\jot]x_{i,2}\\[\jot]x_{i-1,1}\\[\jot]x_{i-1,2}
    \end{pmatrix})=
    \begin{pmatrix}
    x_{i,1}\\[\jot]x_{i,2}\\[\jot]x_{i-1,1}\\\vdots\\[\jot]x_{i,1}x_{i-1,2}\\[\jot]x_{i,2}x_{i-1,2}\\
    \end{pmatrix}=\mathbf{r}_{i+1},
	\label{eq:quadratic_}
\end{equation}
where $\mathbf{r}_{i+1} \in \mathbb{R}^N $ defines the feature vector with feature space dimension $N$.
By concatenating this vector with powers of itself, higher-order features can be included in a computationally cheap way. For this purpose, an additional post-processing operator $\mathbf{q}_{[O_{states}]}(\mathbf{r})$ is introduced, where $O_{states}$ specifies which orders of the feature vector are to be concatenated. Defining $\odot$ as the Hadamard product, $\oplus$ as the vector concatenation operation, and specifying that for $0 \in O_{states}$ a bias term of dimension one is concatenated, the feature space can be extended, for example, for $O_{states}=[0,1,2]$, as shown below,

\begin{align*}
\mathbf{q}_{[0,1,2]}(\mathbf{r}_{i+1}) 
&=1 \oplus \mathbf{r}_{i+1} \oplus (\mathbf{r}_{i+1} \odot \mathbf{r}_{i+1})=(1,r_{i+1,1}, \ldots,r_{i+1,N},r^2_{i+1,1}, \ldots,r^2_{i+1,N})^T=\widetilde{\mathbf{r}}_{i+1} \in \mathbb{R}^{2N+1}
\label{eq:quadratic__}
\end{align*}
where $\widetilde{\mathbf{r}}_{i+1} \in \mathbb{R}^{\widetilde{N}}$ defines the expanded feature vector with dimension $\widetilde{N}$ . This vector is then  mapped with a readout matrix $\mathbf{W}_{out}$ onto the desired output target $\mathbf{y}_{i+1}$. During the training process, this mapping is optimized.
 In the training phase of the NG-RC, the input training data $\mathbf{X}$ of length $T$ is transformed into the feature matrix
 
 \begin{equation}
     \mathbf{R}=\mathbf{q}_{[O_{states}]}(\mathbf{P}^{[O]}(\mathbf{L}^{s}_{k}(\mathbf{X})))
 \end{equation}
 accordingly. Note that due to the $k$ and $s$ value, a warm-up time of $\delta t=ks$ is needed, where entries of the feature matrix at time $t < \delta t$ are not defined. Consequently, the output target matrix $\mathbf{Y}$ needs to be adjusted.
The output target matrix $\mathbf{Y}$ is defined in the scope of this work as

\begin{equation}
\mathbf{Y}=(\Delta \mathbf{x}_{\delta t+1}, \,\ldots \,, \Delta \mathbf{x}_T)^T
\label{eq:DifferenceNG_target}
\end{equation}
with $\Delta \mathbf{x}_i =\mathbf{x}_i-\mathbf{x}_{i-1}$,
 such that the mapping is optimized to fulfill 
 
\begin{equation}
\mathbf{x}_{i+1}=\mathbf{x}_i+\mathbf{W}_{out} \widetilde{\mathbf{r}}_{i+1}.
    \label{eq:differenceNG}
\end{equation}
The readout matrix $\mathbf{W}_{out}$ is learned via ridge regression by optimizing

\begin{equation}
\mathbf{W}_{out}=\mathbf{Y}\mathbf{R}^T(\mathbf{R}\mathbf{R}^T+\beta \mathbf{I})^{-1}.
\end{equation}
Matrix $\mathbf{I}$ is an identity matrix, and $\beta$ is the regression parameter. In this setup, the NG-RC is optimized to become a one-step-ahead integrator that drives the trajectory according to

\begin{equation}
    \mathbf{x}_{i+1}=\mathbf{x}_i+\mathbf{W}_{out} \mathbf{q}_{[O_{states}]}(\mathbf{P}^{[O]}(\mathbf{L}^s_k(\mathbf{x}_i))).\\
\end{equation}

\subsection*{Multifunctionality with input channel}

\subsubsection*{Multifunctionality setup}
To include the data of $n$ trajectories into the training process of the NG-RC, the feature matrix of every trajectory $\mathbf{X}_m$ for $m=1, ...,n$ is calculated with

\begin{equation}
     \mathbf{R}_m=\mathbf{q}_{[O_{states}]}(\mathbf{P}^{[O]}(\mathbf{L}^{s}_{k}(\mathbf{X}_{m}))).
\end{equation}
The resulting feature matrices are concatenated to

\begin{equation}
     \mathbf{R}_M=\mathbf{R}_1 \oplus \mathbf{R}_2 \,\ldots \,  \oplus \mathbf{R}_n.
\end{equation}
The output target matrix for each trajectory $\mathbf{X}_m$ must also be concatenated to

\begin{equation}
     \mathbf{Y}_M=\mathbf{Y}_1 \oplus \mathbf{Y}_2 \,\ldots \,  \oplus \mathbf{Y}_n.
\end{equation}
This way, the identical training routine can be applied so that 

\begin{equation}
\mathbf{W}_{out}=\mathbf{Y}_M\mathbf{R}_M^T(\mathbf{R}_M\mathbf{R}_M^T+\beta \mathbf{I})^{-1}
    \label{muultitrain}
\end{equation}
is optimized via ridge regression.
Provided the training is successful, the $\mathbf{W}_{out}$ can be used to predict the different trajectories 

\begin{equation}
    \mathbf{x}_{m,i+1}=\mathbf{x}_{m,i}+\mathbf{W}_{out} \mathbf{q}_{[O_{states}]}(\mathbf{P}^{[O]}(\mathbf{L}^s_k(\mathbf{x}_{m,i}))).\\
\end{equation}

\subsubsection*{Multifunctionality setup with input channel}

In the scope of this work, however, we use this architecture to modulate the bifurcation parameter for multiple stationary dynamics of a system into the feature vector so that the NG-RC can be tested on predicting the dynamics for new and unseen bifurcation parameters.
Therefore, for every stationary dynamic $\mathbf{X}_m$ in the training data, determined by its bifurcation parameter $\theta_m$, we add to each element in the corresponding feature vector the bifurcation parameter $\theta_m$ multiplied by a scaling parameter $\gamma$,

\begin{equation}
     \mathbf{R}_m=\mathbf{q}_{[O_{states}]}(\mathbf{P}^{[O]}(\mathbf{L}^{s}_{k}(\mathbf{X}_{m}))+\gamma \theta_m).
\end{equation}
This concept can be trained similarly by optimizing Eq. \ref{muultitrain}. The NG-RC then drives the trajectory according to

\begin{equation}
    \mathbf{x}_{i+1}=\mathbf{x}_i+\mathbf{W}_{out} \mathbf{q}_{[O_{states}]}(\mathbf{P}^{[O]}(\mathbf{L}^s_k(\mathbf{x}_i))+\gamma \theta).\\
\end{equation}
In addition, this structure allows to change the bifurcation parameter per prediction step,

\begin{equation}\
    \mathbf{x}_{i+1}=\mathbf{x}_i+\mathbf{W}_{out} \mathbf{q}_{[O_{states}]}(\mathbf{P}^{[O]}(\mathbf{L}^s_k(\mathbf{x}_i))+\gamma \theta_i).\\
    \label{non-stationary equation}
\end{equation}
Provided that the trained architecture can predict the dynamical behavior of the system for various unseen bifurcation parameters successfully, i.e., reconstruct its bifurcation diagram, a reasonable motivation for simulating non-stationary processes can be derived from Eq. \ref{non-stationary equation}.

\subsection*{Lyapunov exponent}

Due to the definition of chaos and its sensitivity to initial conditions, evaluating dynamical predictions only on their deviation from the ground truth, i.e., with its short-time behavior, can not capture essential features of dynamical systems. To determine the systematic behavior of a dynamical system, it is necessary to determine the long-term properties of the trajectory. These are referred to as the statistical climate of the system, and its measurement can give rise to how chaotic or periodic the system is. Lyapunov exponents $\lambda_{i}$ measure the temporal complexity of the dynamical system by measuring the average divergence rate of nearby points in phase space. This gives rise to its sensitivity to initial conditions for each dimension $i$ and quantifies the time scale on which it becomes unpredictable \cite{wolf1985determining, shaw1981strange}. Suppose at least one Lyapunov exponent is positive. In that case, the system is considered chaotic. The magnitude of the largest Lyapunov exponent $\lambda_{max}$ can then be taken to measure the degree of chaoticity the system exhibits. In the context of this work, the Rosenstein algorithm is used to calculate the largest Lyapunov exponent 
\cite{rosenstein1993practical}.

\subsection*{Power System Model}
\subsubsection*{Model equations}
The model consists of four ordinary differential equations.

\begin{equation}
    \dot{\delta}_m=\omega,
\end{equation}
\begin{equation}
       M\dot{\omega}=-d_{m}\omega+P_m-E_mY_msin(\delta_m-\delta)V , 
\end{equation}
\begin{equation}
    K_{qw}\dot{\delta}=-K_{qv2}V^2-K_{qv}V+Q(\delta_m,\delta,V)-Q_0-Q_1 ,
\end{equation}
\begin{align}
    TK_{qw}K_{pv}\dot{V}=& K_{pw}K_{qv2}V^2 +(K_{pw}K_{qv}-K_{qw}K_{pv})V +K_{qw}[P(\delta_m,\delta,V)-P_0-P_1]
    -K_{pw}[Q(\delta_m,\delta,V)-Q_0-Q_1] 
\end{align}
where

\begin{align}
       P(\delta_m,\delta,V)=&-E'_0Y'_0Vsin(\delta)+E_mY_mVsin(\delta_m-\delta) \nonumber, 
\end{align}
\begin{align}
       Q(\delta_m,\delta,V)=&-E'_0Y'_0Vcos(\delta) -(Y'_0+Y_m)V^2+E_mY_mVcos(\delta_m-\delta). \nonumber
\end{align}
The real power demand $P$ and the reactive power demand $Q$ of the system appear in the differential equations of the load voltage $V$ and the motor frequency $\delta$. The variable ${\delta}_m$ describes the angle dynamics between two generators and $\omega$ the speed of a generator rotor. For a more technical description, we suggest the paper by Dobson et al. \cite{dobson1989towards}.
Following the model parameterization
used for the traditional reservoir computing approach\cite{kong2021machine},

\begin{equation}
    E'_0=\frac{E_0}{(1+C^2Y_0^{-2}-2CY_0^{-1}cos(\theta_0))^{\frac{1}{2}}} ,
\end{equation}
\begin{equation}
    Y'_0=Y_0(1+C^2Y_0^{-2}-2CY_0^{-1}cos(\theta_0))^{\frac{1}{2}} ,
\end{equation}
\begin{equation}
    \theta'_0=\theta_0+tan^{-1} \left (\frac{CY_0^{-1}sin(\theta_0)}{1-CY_0^{-1}cos(\theta_0)}\right)
\end{equation}
are set as constants and $K_{pw}=0.4$, $K_{pv}=0.3$, $K_{qw}=-0.03$, $K_{qv}=-2.8$, $K_{qv2}=2.1$, $T=8.5$, $P_0=0.6$, $Q_0=0.3$, $P_1=0$, $Y_0=3.33$, $Y_m=5$, $P_m=1$,
$d_m=0.05$, $\theta_0=0$, $E_m=1.05$, $M=0.01464$, $C=3.5$, $E_0=1$, $Q_0=1.3$.

$Q_1$ is taken as the bifurcation parameter. It determines the load reactive power demand of the system. The model bifurcation diagram was created using Runge-Kutta 4 (RK4), starting each time from $\mathbf{x}_0=(\delta_{m,0},\omega_0,\delta_0,V_0)^T=(0.17,0.05,0.05,0.83)^T$ for $T=10000$ time steps with time step size $\Delta t=0.05$ and an bifurcation parameter step size of $\Delta Q_1=0.000001$.

\subsubsection*{NG-RC architecture}

The training data is generated identically for each system parameter in 

\begin{align}
    \mathbf{Q}^{train}_1=\, &[2.98953, 2.98956, 2.98960, 2.98964, 2.98967, 2.98969, 2.98975]. \nonumber
\end{align}
The following NG-RC architecture

\begin{equation}
     \mathbf{R}_{Q_{1,m}}=\mathbf{q}_{[0,1,2,3]}(\mathbf{P}^{[1,2,3]}(\mathbf{L}^{2}_{2}(\mathbf{X}_{m}))+\gamma Q^{train}_{1,m}),
     \label{PSM_good}
\end{equation}
is used and trained with a regression parameter of $\beta=10^{-8}$. The expanded feature vectors have dimension  $\widetilde{N}=493$.
The warm-up times of length $\delta t=4$ are simulated with RK4.

\subsection*{Chaotic Food Chain Model}
\subsubsection*{Model equations}
The three-species food chain model with a resource density $R$,
a consumer density $C$, and a predator density $P$ is modeled as follows 

\begin{equation}
    \dot{R}=R\left(1-\frac{R}{K}\right)-\frac{x_cy_cCR}{R+R_0},
\end{equation}

\begin{equation}
    \dot{C}=x_cC\left(\frac{y_cR}{R+R_0}-1\right)-\frac{x_py_pPC}{C+C_0},
\end{equation}

\begin{equation}
    \dot{P}=x_pP\left(\frac{y_pC}{C+C_0}-1\right)
\end{equation}
with $x_c=0.4$, $y_c=2.009$, $x_p=0.08$, $y_p=2.876$, $R_0=0.16129$, $C_0=0.5$ and the resource-carrying capacity $K$ is taken as the bifurcation parameter\cite{kong2021machine}.
The model bifurcation diagram was created using RK4, starting each time from $\mathbf{x}_0=(R_0,C_0,P_0)^T=(0.6,0.35,0.9)^T$ for $T=25000$ time steps with time step size $\Delta t=0.1$ and an bifurcation parameter step size of $\Delta K=0.00025$.

\subsubsection*{NG-RC architecture}
The training data is generated identically for each parameter in 

\begin{align}
    \mathbf{K}^{train}=[0.92, 0.925, 0.93, 0.935, 0.94, 0.945, 0.95]. \nonumber
\end{align}
The following NG-RC architecture

\begin{equation}
     \mathbf{R}_{K_{m}}=\mathbf{q}_{[0,1,2,3]}(\mathbf{P}^{[1,2]}(\mathbf{L}^{4}_{4}(\mathbf{X}_{m}))+\gamma K^{train}_{m})
     \label{CFC_good}
\end{equation}
is applied. The expanded feature vectors have dimension $\widetilde{N}=271$. A regression parameter of $\beta=10^{-3}$ is used. The warm-up times of length $\delta t=16$ are simulated with RK4.

\bibliography{sample}

\section*{Acknowledgements}

D.K. gratefully acknowledges the funding provided by Allianz Global Investors (AGI).

\section*{Author contributions statement}

C.R. initiated the research. D.K. and C.R. designed the study. D.K. conducted the calculations. C.R. and D.K. interpreted and evaluated the findings. All authors reviewed the manuscript. 

\section*{Competing interests}
All authors declare no competing interests.

\section*{Data Availability}

The datasets used and/or analysed during the current study available from the corresponding author on reasonable request.

\end{document}